# Position Focused Attention Network for Image-Text Matching

**Yaxiong Wang**[1,2*], **Hao Yang**[1†], **Xueming Qian**[2], **Lin Ma**[3], **Jing Lu**[1], **Biao Li**[1] and **Xin Fan**[1]

[1]Department of PCG, Tencent
[2]School of Software Engineering, Xi'an Jiaotong University, China
[3]Tencent AI Lab
wangyx15@stu.xjtu.edu.cn, applehyang@tencent.com
qianxm@mail.xjtu.edu.cn, forest.linma@gmail.com

## Abstract

Image-text matching tasks have recently attracted a lot of attention in the computer vision field. The key point of this cross-domain problem is how to accurately measure the similarity between the visual and the textual contents, which demands a fine understanding of both modalities. In this paper, we propose a novel position focused attention network (PFAN) to investigate the relation between the visual and the textual views. In this work, we integrate the object position clue to enhance the visual-text joint-embedding learning. We first split the images into blocks, by which we infer the relative position of region in the image. Then, an attention mechanism is proposed to model the relations between the image region and blocks and generate the valuable position feature, which will be further utilized to enhance the region expression and model a more reliable relationship between the visual image and the textual sentence. Experiments on the popular datasets Flickr30K and MS-COCO show the effectiveness of the proposed method. Besides the public datasets, we also conduct experiments on our collected practical large-scale news dataset (Tencent-News) to validate the practical application value of proposed method. As far as we know, this is the first attempt to test the performance on the practical application. Our method achieves the state-of-art performance on all of these three datasets.

## 1 Introduction

In recent years, the computer version has developed rapidly, the focused tasks are moving from predicting the object categorical labels to more challenging problems. Image-text matching is one of the important branches for various visual related applications, such as bi-directional image and text

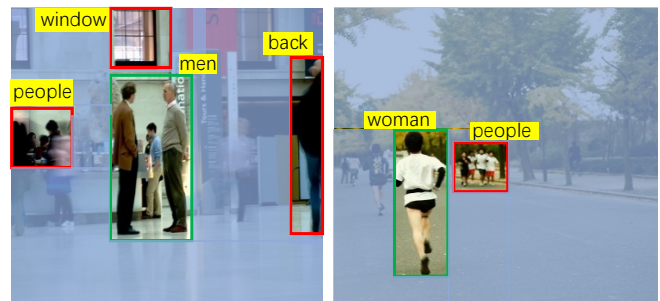

(a) Two men in formal wear talking next to people moving

(b) A woman in a white top runs along a tree lined path

Figure 1: Position can indicate the importance of the regions

retrieval [Yan *et al.* 2015; Ma *et al.* 2015], natural language object retrieval [Hu *et al.*, 2016], image captioning [Xu *et al.*, 2015; Vinyals *et al.*, 2017], and visual question answering (VQA) [Antol *et al.*, 2015; Lin *et al.*, 2016]. Therefore, many researchers have dedicated their efforts to study the relationship between the visual and the textual contents [Lee *et al.*, 2018; Gu *et al.*, 2018; Eisenschtat *et al.*, 2017; Ma *et al.*, 2015; Wang *et al.*, 2018; Klein *et al.*, 2015].

Image and text both contain rich information but reside in heterogeneous modalities. Comparing to information retrieval within the same modality, the designed model for cross-modal retrieval need not only learn the features for image and text to express their respective content but a measure for cross-modal similarity calculation. Therefore, cross-modal retrieval poses extra critical challenges. A popular framework to model the relationship between image and text is the two-branch embedding network, one branch projects the image and another models the text, the shared subspace is learned by the popular triplet loss [Gu *et al.*, 2018; Wang *et al.*, 2018; Nam *et al.*, 2017]. Besides the network structure, recently, more and more scholars design their embedding networks based on attention mechanism, which attempts to capture the correspondences between the detected visual objects and the textual items (words or phrases). Many studies have validated that the attention is helpful to model a more reliable relationship between image and text [Lee *et al.*, 2018; Nam *et al.,* 2017; Huang *et al*., 2017; Anderson *et al*.,

2018].

However, existing attention mechanisms only focus on the visual feature of the image regions while ignore the relative position information in the images. In general, if an object region is closer to the center, it may express the main semantics of the image with higher probability, while the marginal ones may not be that important. Just as shown in Figure 1(a), the main semantic part corresponding to the word "men" locates at the center of the picture, while the peddling objects lie on the brink. From this observation, an intuitive idea is to simply pay more attention to the regions closer to the center. However, not all regions near the center are that important, as shown in Figure 1(b) which exhibits that a woman (the most important object) lies in the lower left part. Furthermore, simply assigning attention to the region based on the fixed position (the center for example) cause a bad extendibility. From above observations and considerations, we design a position feature for the region to integrate position information and propose an attention mechanism to generate the valuable position feature for each region.

In this paper, a novel position focused attention network is developed to study the fine-grained interplay between the image regions and the words. Our contributions can be summarized as follows:

1) A novel position feature is designed for image regions, by which we integrate the position information of regions to investigate the correspondences between objects in image and words in sentence.

2) We propose a position focused attention mechanism to generate the valuable position feature for the image region, and the position feature together with the visual feature form a more reliable and complete expression of the image region.

3) Besides two public datasets, we make the first attempt to evaluate the application value on a practical news dataset and our method achieves the state-of-art performance on all of these three datasets.

The remainder of this paper is organized as follows: In section 2, we elaborate the details of each process in our system. Experiments are shown in section 3. Finally, conclusions and future works are given in section 4.

## 2 Our Approach

In this section, we will elaborate the details of our proposed framework. Figure 2 shows the flowchart of this paper, we first extract the features of the region and the position, the visual feature together with the generated position feature form the final region's representation, and the alignments between the region and the word are studied by a visual-textual attention [Lee *et al.*, 2018]. The network is trained by the triplet ranking loss.

Next, we describe the input representation in subsection 2.1. In subsection 2.2, the position information integration is presented, subsection 2.3 presents the image-sentence relevance calculation.

### 2.1 Input Feature Representation

**Image feature.** In this paper, an image $I$ is represented by a set of features $\{v_1, v_2, ..., v_n\}$, and $v_i \in R^D$, where $n$ is the number of image regions. Since our attention mechanism is focused on the image regions, especially the objects in image. Therefore, we detect the objects in image utilizing the Faster R-CNN model [Ren *et al.*, 2017]. In order to get a better feature representation, we feed the detected object into the ResNet-101 [He *et al.*, 2016] pre-trained on Visual Genomes [Krishna *et al.*, 2017] by Anderson *et al.* [Anderson *et al.*, 2018] to extract the visual feature. Finally, the input image is represented by $n$ $D$-dimensional feature vectors, and $D$ is 2048 in our experiment.

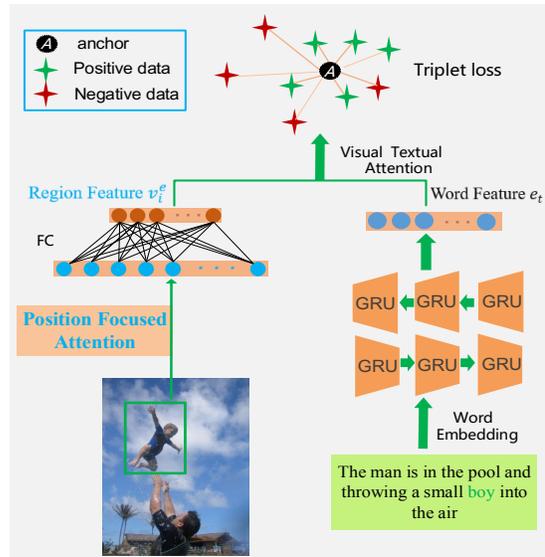

Figure 2: The flowchart of proposed network

**Text feature.** On the subject of corresponding image description, the basic item is the word in the sentence. Each word is represented with one-hot vector, which indicates the index in the vocabulary. Then the one-hot representation is embedded into $d$-dimensional vector by a linear mapping layer, $x_t = W \times w_t$, where $w_t$ is a one-hot of word in a sentence with $T$ words $\{w_1, w_2, ..., w_T\}$, $W \in R^{d \times N}$ is the embedding matrix, $N$ is the vocabulary size.

### 2.2 Position Information Integration

The relative position of the object in the whole image is an important and useful clue, which is helpful to infer the significance of the object region, just as shown in Figure 1. Motivated by this observation, we fuse the position information into the learning procedure to capture more reliable and credible fine-grained interplay between the image and the text elements. In this subsection, we present our position attention mechanism. We first introduce the initial positional representation, and then elaborate the block embedding. Finally, our position focused attention is presented.

**Initial Position Representation**

Given an image $I = \{v_1, v_2, ..., v_n\}$, in order to reveal the relative position for a region $v_i$ in the whole image $I$, we first equally split the whole image $I$ into $K \times K$ blocks $B$, the position of each block is initially represented by an index $k \in [1, K^2]$, we locate the region $v_i$ according to its overlap with these fixed blocks. Let $p_i \in R^L$ denote the position index vector of region $v_i$, which is defined as the indexes of the top $L$ max overlapping blocks with the region $v_i$, i.e. the indexes in $p_i$ meet:

$$OV(v_i, b_{p_{ij}}) \geq OV(v_i, b_q), \ j = 1, 2, \ldots, L \quad (1)$$

where $p_{ij} \in [1, K^2]$ is the block index of the $j$-th maximum overlapping with the region $v_i$, $q \in [1, K^2] \setminus p_i$, the operator "\" means removing, and $OV(v_i, b_q)$ is the intersecting pixel number between region $v_i$ and the $q$-th block:

$$OV(v_i, b_q) = |v_i \cap b_q| \quad (2)$$

where $b_q \in B$. We also define an additional vector $a_i \in R^L$ for region $v_i$ to record the corresponding overlapping to distinguish the importance of different blocks:

$$a_{ij} = OV(v_i, b_{p_{ij}}) \in R \quad (3)$$

and $a_{ij}$ is normalized for further processing.

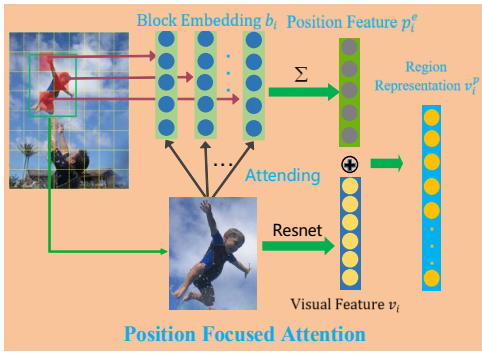

Figure 3: The proposed position focused attention mechanism

**Block Embedding**

$L$ indexes of blocks are introduced to indicate the relative position of the region. To get a more accurate description for the position, we embed the block index into a dense representation. The split blocks $B$ are regarded as the position vocabulary, and each block $b_i \in B$ is represented by the one-hot vector, which indicates the index in the position vocabulary. We next apply an embedding layer to project the one-hot representation into $\iota$-dimensional vector, we still denote the new embedding vector as $b_i$ for the sake of simplicity.

We can then simply generate the position representation of region $v_i$ based on the embedding block vector:

$$p_i^e = \sum_{j=1}^L b_{p_{ij}} \times a_{ij} \quad (4)$$

**Position Focused Attention**

After obtaining the block embedding, we can represent the position of region $v_i$ according to the Eq. (4). However, it's insufficient to directly use the rate of the overlapping area. Since there are many blocks completely covered by the region and the contributions of these blocks will be equal accordingly. From this consideration, an adaptive weight is assigned to each block with respect to each region. As shown in Figure 3, the proposed attention aims at deciding how much weight should be paid to the position cell for the current region:

$$\beta_{ij} = \tanh(f(v_i, b_{p_{ij}})), i \in [1, n], j \in [1, L] \quad (5)$$

where $f$ is the bilinear function:

$$f(v_i, b_{p_{ij}}) = v_i^T M b_{p_{ij}} \quad (6)$$

where $M \in R^{D \times \iota}$ is the mapping matrix.

Besides the completely covered blocks should be different, another intuition is that the more of the block is covered by the region, the more important it should be. According to above considerations, we improve the Eq. (4) accordingly as following:

$$p_i^e = \sum_{j=1}^L b_{p_{ij}} \times \gamma_{ij} \quad (7)$$

and

$$\gamma_{ij} = \frac{\gamma'_{ij}}{\sum_j \gamma'_{ij}}, \text{where } \gamma'_{ij} = \frac{\exp(\beta_{ij})}{\sum_j \exp(\beta_{ij})} \times a_{ij} \quad (8)$$

The final position representation of region $p_i^e$ is then concatenated with the visual feature $v_i$, such that the region feature will carry position information, i.e. $v_i^p = [v_i, p_i^e] \in R^{D+\iota}$.

### 2.3 Image-Sentence Relevance Calculation

Given an image $I$ with $n$ regions and a sentence $S$ with $T$ words, we utilize a fully-connected layer to project the final region representation $v_i^p$ into a $h$-dimensional feature $v_i^e \in R^h$. As for the words, the final feature is obtained by feeding the embedding vector into a bi-directional GRU [Zhu et al., 2017], whose dimension of the hidden state is also set as $h$. The final representation $e_t$ of the word is the average of forward and backward feature:

$$e_t = \frac{e_t^f + e_t^b}{2} \in R^h \quad (9)$$

where $e_t^f$ and $e_t^b$ are the forward and backward features respectively.

Following the work in [Lee et al., 2018], an attention weight $\alpha_{it}$ for region $v_i$ with respect to the word $w_t$ is calculated, which decides how much attention to pay the region for current word. The visual vector of the current word is then defined as the weighted combination of region representation:

$$\nu_t = \sum_{i=1}^n \alpha_{it} v_i^e \quad (10)$$

The semantic relevance between the image $I$ and the sentence $S$ is taking the average of the relevance between all the semantic features in $S$ and the attending visual vectors:

$$S(I, S) = \frac{\sum_t r(e_t, \nu_t)}{T} \quad (11)$$

where $T$ is the number of words in the sentence, $r(\cdot, \cdot)$ is the cosine similarity function.

On the other hand, the procedure of attending image to the text is analogous to the above. The triplet loss with the hardest negative sample is employed for network training, which is a common ranking objective for image-text matching. The training triplets are sampled within a mini-batch by the sampling strategy in [Lee *et al.*, 2018; Faghri *et al.*, 2018].

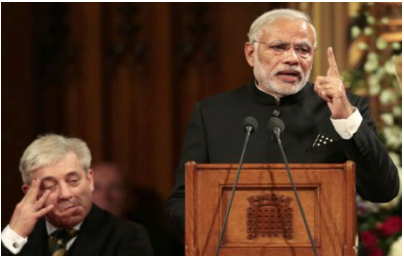

Figure 4: An exemplar image of Tencent-News with title "莫迪希望在安保及经济方面强化与日本战略关系" (Modi hopes to strengthen strategic relationship with Japan in security and economy)

## 3 Experiments

In order to demonstrate the effectiveness of our proposed methods, we conduct our Position Focused Attention Network (PFAN) on two public datasets: Flickr30K and MS-COCO, and a practical Chinese news dataset: **Tencent-News**[1]. We systematically make comparisons with several latest start-of-the-art methods and thoroughly investigate the performance of the proposed PFAN. As for the performance measure criterion for sentence retrieval or image retrieval, we apply the commonly used recall on top H (R@H), which is defined as the percentage of correct items in the top H retrieved results.

### 3.1 Implement Details

**Dataset.** We evaluate our PFAN on the widely used and authoritative dataset Flickr30K, MS-COCO, the data splits for these two datasets follow the work [Karpathy *et al.*, 2015] and [Lee *et al.*, 2018]. Besides the public datasets, a practical news dataset, Tencent-News, is also collected to evaluate the value of the proposed method in the practical application. This dataset can be used for training image-text models to further support Chinese corpus.

**Tencent-News.** For a piece of news, the title and one perfect matching image in this news make up a basic data item, and the title is regarded as the description of this image. By this way, we collect 143,317 training pairs, and 1,000 pairs for validating and there are 141,736 different images, 130,230 different titles in total. In the test procedure, we manually label 510 news and several corresponding images ($\geq 5$) for performance evaluation. There are 510 titles and 2,794 labeled images in total. Each title has 5.5 candidate images, 2.3 irrelevant images and 3.2 relevant images on average. Figure 4 shows an exemplar image of this dataset. In this practical dataset, we focus on the news auto-image recommendation task, i.e. the news editor inputs the news title, and the model can automatically output several related candidate images for this news, which can remarkably alleviate the effort of editors and speed up the news publish. In this application scene, we only focus on the task of the image retrieval.

**Training details.** All of our experiments are conducted on a workstation with NVIDIA Tesla GPU. Adam optimization algorithm is used to train the overall network. The mini-batch size is 128. The image region is extracted by the Faster R-CNN model [Ren *et al.*, 2017], and we retain 36 detected regions for the image representation. Each image is split into 16×16 blocks ($K$=16), and we set $L$ as 15. The dimension of joint embedding is fixed as 1024. The block index is first embedded into 200-dimensional space, and the original 2048-dimensional visual vector together with 200-dimensional position feature is mapped into the 1024-dimensional space by a linear projection layer. On the subject of word, the one-hot vector is first embedded into 300-dimensional dense representation, then the dense representation is fed into the bi-GRU whose hidden dimension is set as 1024 as well. For Flickr30k and MS-COCO dataset, the training details are the same as reference [Lee *et al.*, 2018]. On the Tencent-News dataset, the parameter settings except the embedding size are the same as the Flickr30K, we set the embedding size as 512 to get better performance.

### 3.2 Performance Evaluation

**Comparison with the Competing Methods**

Table 1 shows the performances of all methods on Flickr30k dataset, where PFAN t-i means only employing the loss of attending text to image to train the network and PFAN t-i+i-t fuses the models from PFAN t-i and PFAN i-t. As shown in Table 1, our PFAN t-i and PFAN i-t both achieve satisfied performance. For example, when the depth varies from 1 to 10, the recall of PFAN t-i can reach 66.0, 89.6 and 94.3 respectively on the sentence retrieval given an image. The fused model PFAN t-i+i-t makes our performance step further and outperforms the current state-of-art method SCAN (i-t+t-i fused), the R@1 on sentence retrieval task can even reach 70.0, which validates the superior of our proposed method.

Table 2 exhibits the performances on the MS-COCO dataset, from which we can also observe that PFAN is more outstanding than other methods and the similar conclusions can be obtained.

Table 3 shows the performances on Tencent-News dataset. In order to evaluate the retrieved images by a given news title, we use the Mean Average Precision (MAP) [Qian et al., 2017; Wang *et al.*, 2018] and the Accuracy (A) to evaluate the performance. It is clear from Table 3 that our PFAN is still more outstanding than the state-of-art method SCAN, PFAN can outperform SCAN by six points on average under both MAP and Accuracy. Tables 1-3 show the effectiveness and the practical applied value of the proposed method.

### 3.3 Result Visualization

In this subsection, we will visualize some results to see the ability of the proposed network on relationship capture aspect.

---

[1] The Tencent News data download link and our code can be found at: https://github.com/HaoYang0123/Position-Focused-Attention-Network/

| methods | Image-to-Text Retrieval | | | Text-to-Image Retrieval | | |
|---|---|---|---|---|---|---|
| | R@1 | R@5 | R@10 | R@1 | R@5 | R@10 |
| SM-LSTM[Huang et al., 2017] | 42.5 | 71.9 | 81.5 | 30.2 | 60.4 | 72.3 |
| 2WayNet [Eisenschtat et al., 2017] | 49.8 | 67.5 | - | 36.0 | 55.6 | - |
| DAN [Nam et al., 2017] | 55.0 | 81.8 | 89.0 | 39.4 | 69.2 | 79.1 |
| VSE++ [Faghri et al., 2018] | 52.9 | - | 87.2 | 39.6 | - | 79.5 |
| DPC [Zheng et al., 2018] | 55.6 | 81.9 | 89.5 | 39.1 | 69.2 | 80.9 |
| SCO [Huang et al., 2018] | 55.5 | 82.0 | 89.3 | 41.1 | 70.5 | 80.1 |
| SCAN [Lee et al., 2018] | 67.4 | 90.3 | **95.8** | 48.6 | 77.7 | 85.2 |
| PFAN t-i | 66.0 | 89.6 | 94.3 | 49.6 | 77.0 | 84.2 |
| PFAN i-t | 67.6 | 90.0 | 93.8 | 45.7 | 74.7 | 83.6 |
| PFAN t-i+i-t | **70.0** | **91.8** | 95.0 | **50.4** | **78.7** | **86.1** |

Table 1: Comparisons of cross-modal retrieval on Flickr30K dataset with the competing methods

| methods | Image-to-Text Retrieval | | | Text-to-Image Retrieval | | |
|---|---|---|---|---|---|---|
| | R@1 | R@5 | R@10 | R@1 | R@5 | R@10 |
| SM-LSTM[Huang et al., 2017] | 53.2 | 83.1 | 91.5 | 40.7 | 75.8 | 87.4 |
| 2WayNet [Eisenschtat et al., 2017] | 55.8 | 75.2 | - | 39.7 | 66.3 | - |
| DAN [Nam et al., 2017] | 55.0 | 81.8 | 89.0 | 39.4 | 69.2 | 79.1 |
| VSE++ [Faghri et al., 2018] | 64.6 | - | 95.7 | 52.0 | - | 92.0 |
| DPC [Zheng et al., 2018] | 65.6 | 89.8 | 95.5 | 47.1 | 79.9 | 90.0 |
| SCO [Huang et al., 2018] | 69.9 | 92.9 | 97.5 | 56.7 | 87.5 | 94.8 |
| SCAN [Lee et al., 2018] | 72.7 | 94.8 | 98.4 | 58.8 | 88.4 | 94.8 |
| PFAN t-i | 75.8 | 95.9 | 99.0 | 61.0 | 89.1 | 95.1 |
| PFAN i-t | 70.7 | 94.1 | 97.8 | 53.0 | 84.5 | 92.6 |
| PFAN t-i+i-t | **76.5** | **96.3** | **99.0** | **61.6** | **89.6** | **95.2** |

Table 2: Comparisons of cross-modal retrieval on MS-COCO dataset with the competing methods

| | MAP@1 | MAP@3 | MAP@3 | A@1 | A@2 | A@3 |
|---|---|---|---|---|---|---|
| SCAN | 67.2 | 70.6 | 75.7 | 67.2 | 69.1 | 73.6 |
| PFAN | **76.0** | **79.0** | **82.0** | **76.0** | **76.3** | **79.7** |

Table 3: Performances on Tencent-News dataset

**Position Embedding Visualization**
In this work, we employ an embedding layer to project the block index into a dense representation. Intuitively, the learned block index embedding should preserve the locality, i.e. the neighbor position embeddings should be close to each other. In order to see if the learned position embedding preserve the locality, we compute the similarity matrix $SM = R^{16\times16}$ of the block position embeddings, and the component $SM(i,j)$ is defined as the average similarity between the block embedding in $i$-th row and $j$-th column and its adjacent embeddings:

$$SM(i,j) = \frac{1}{|\mathcal{L}(b_m)|}\sum_{d\in\mathcal{L}(b_m)} \exp\left(-\frac{\|b_m-d\|^2}{2\sigma^2}\right) \quad (12)$$

the $SM$ is calculated based on Gaussian kernel function, where the $m = i\times 16 + j$, $\mathcal{L}(b_m)$ is the collection of the position embeddings that are directly adjacent to the $b_m$, $\sigma \in R$ is a scalar, we simply set it as the average distance between all the position embeddings.

Figure 5 shows the visualization result of matrix $SM$, we can find that the closer block position embeddings share higher similarity in most cases. For example, the four blocks marked as (a)-(d) are very similar to each other, and many analogy situations can be found in Figure 5, which means the learned position embeddings preserve the locality.

We can also observe that the similarities of some position embeddings close to the center are with lower value ((c), (d)), while many marginal blocks are highly similar to each other.

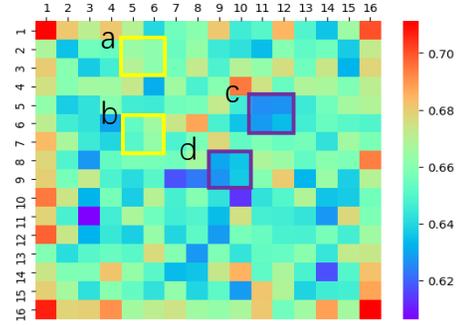

Figure 5: The visualization of position embedding similarity

We guess that this is because the regions in the corners are relatively similar, just as shown in the lower right region in Figure 1(a) and (b). Therefore, the position features in corner frequently meet similar visual content and fit the similar visual content in each iteration. The final learned position embedding should be similar. However, the content close to the center in different images are violently changed, the corresponding block position needs to fit various visual contents, therefore, there will be obvious differences accordingly.

**Attention Visualization**
We design a position attention mechanism to adaptively determine the importance of the block position according to the attention weights instead of the fixed position. More details can be found in section 2. In this subsection, we visualize the attention results.

An exemplary visualization result is shown in Figure 6, where the green box indicates the image region, the word with

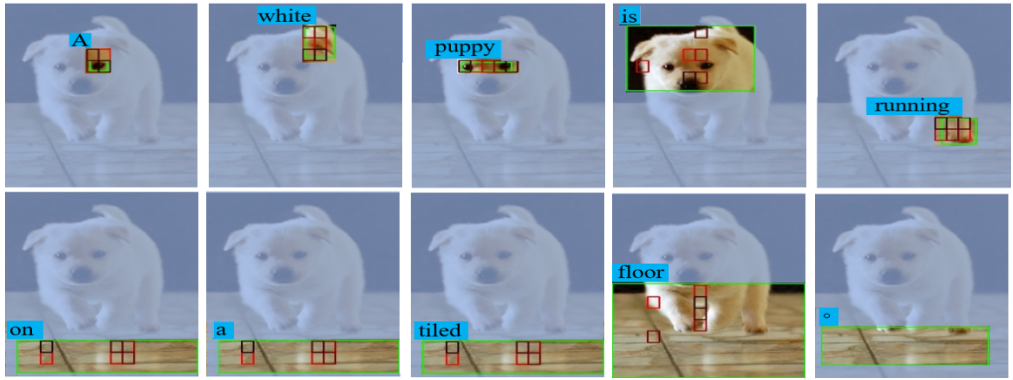
Figure. 6: The visualization figures of attending image region to each word

the maximum attention weight to the region is exhibited in each figure. The red frames indicate the blocks that the region attending, we exhibit the blocks with the first 6 maximum weights for each region and the brighter ones are with higher weights. There are two observations from Figure 6:

a. The correspondences between the image regions and the sentence words are satisfied, most of the words can attend to their related semantic regions, like the words "white", "running", "floor", and so on.

b. The brighter blocks indeed reveal more important parts of the regions. Take the fourth image in the first row as an example, the brightest block locates in the center of the region. From the fifth image, we can get the similar observation.

## 4  Conclusion and Future Work

In this paper, we develop a position focused attention network (PFAN) for the image-text matching task. Instead of only paying attention to the regions themselves, the clue of the region position is taken into consideration. We first split the image and utilize the split blocks to infer the relative position of the region, an attention weight is then assigned to the block with respect to each region and position feature is then adaptively generated by the designed position attention. Positional feature and visual feature are concatenated to form the final representation of the region. The experiments on the popular Flickr30k and MS-COCO datasets reveal that integrating the position information can help model a more reliable relation between the image and the text. Besides the public datasets, we further collect a practical dataset (Tencent-News) and make the first attempt to evaluate the application value of our image-text model. The results on these three datasets are all much better than the competing methods and achieve the state-of-art performance. In the future, we will fuse more semantic information to enhance the cross-domain relation learning.

## Acknowledgements

This work was supported in part by the NSFC under Grant 61772407 and 61732008.